# The Analysis of Facial Feature Deformation using Optical Flow Algorithm

Dayang Nur Zulhijah Awang Jesemi, Hamimah Ujir, Irwandi Hipiny and
Sarah Flora Samson Juan
Universiti Malaysia Sarawak, Malaysia.

**Abstract**

Facial features deformed according to the intended facial expression. Specific facial features are associated with specific facial expression, i.e. happy means the deformation of mouth. This paper presents the study of facial feature deformation for each facial expression by using an optical flow algorithm and segmented into three different regions of interest. The deformation of facial features shows the relation between facial the and facial expression. Based on the experiments, the deformations of eye and mouth are significant in all expressions except happy. For happy expression, cheeks and mouths are the significant regions. This work also suggests that different facial features' intensity varies in the way that they contribute to the recognition of the different facial expression intensity. The maximum magnitude across all expressions is shown by the mouth for surprise expression which is $9 \times 10^{-4}$. While the minimum magnitude is shown by the mouth for angry expression which is $0.4 \times 10^{-4}$.

Keywords: Facial expression, Facial features, Facial deformation and Optical Flow.

1. Introduction

   Facial expression is one of the responses to human emotional stimuli. Recognizing facial expression is one of the ongoing works in face processing area of study. Face processing brings benefits to several fields such as security, medicine, and marketing. In the security field, facial expression contributes to detection of the crowd's mood in a controlled environment. For medicine area, facial expression recognition could help to assess the pain and identify the depression state of a patient during treatment. While in marketing, by knowing the current mood of the crowd in public places, certain measurement can be taken to improve the crowd's mood and a different marketing strategy can be offered by the expert. It brings benefits in human centred multimodal human-computer interaction (HCI) where the affective computing in HCI domain employs human emotion to build more flexible and natural multimodal [1].

   Facial features are the important features which are needed as the input in the development of a human-machine interaction, specifically for facial based technology. In face processing area, facial features such as eye, nose, mouth, cheeks and eyebrows are related to the facial expression interaction. Facial features will be deformed according to the intended facial expression. By tracking the facial features when a facial expression occurred, the dynamic of the emotion shown can be understood. Although the facial expression changes can be observed by the human eye, the movement of the facial feature is hard to observe. Thus, the intention of this study is to identify which facial features are deformed when these facial expressions occurred. Besides, the level of the facial expression intensity could be identified. Expression can be measured from low to a high level based on the sequence of expression through facial feature deformation. By detecting the movement of facial features, the intensity of the emotional state can be measured.

   In this paper, the optical flow algorithm used to track facial feature deformation for six basic facial expressions (angry, disgust, fear, happy, sad, surprise). The intensity of facial expression is also measured. Section 2 describes the research method used in this study and followed by a discussion of results and analysis in Section 3. Finally, the conclusions are drawn.

## 2. Related Work

Facial expression study attracts researchers in artificial intelligent, computer vision and several other fields of studies, which includes the psychologist. Different researchers used different facial features in face processing studies. [18] proposed 7 most expressive facial regions excluding the eyes and mouth. Facial distance vectors are used in 3D facial expression analysis and classification [19][20][21][25].

In [16], participants were asked to identify expressions from female and male faces displaying six basic universal expressions (anger, disgust, fear, happiness, sadness, and surprise), each with three levels of intensity (low, moderate, and normal). Only two facial regions are considered in [1] which is the eyes and mouth region. One of their findings is the mouth region was the most significant for fear, angry, and disgust expressions and least for surprise. Expression changes models are developed by [2] and they form distinctive quantitative profiles of expressions, known as a Regional Volumetric Difference (RVD) function. Expression classifiers for the four-universal facial expression, anger, fear, happy and sad, are designed by training on RVD functions of expression changes.

In another domain, [3] proposed a framework based on Dynamic Bayesian Network (DBN) to systematically model relationships among spontaneous AUs for measuring their intensities. [4] analyze facial expressions with four intensities in 3D space by exploring the facial distances. The extracted distance measurements provide information for the facial expression intensity measurement as well as the classification of facial expressions. This work reveals that it is not necessary to rely on all facial feature points in estimating facial expression intensity.

According to [5], the decomposition of a face into several modules promotes the learning of a facial local structure and therefore the most discriminative variation of the facial features in each module is emphasized. There are several studies that employed face decomposition in their work [15][22][23][24]. The work in [4] considers three face regions while work in [16] only measured two regions. According to [5], each face region has its own impact level to the six basic expressions and the simplest way to find which modules are affecting facial expression is to find their priority weighting. [5] computed the weighting of each face region using Adaboost. Works in [15] subdivided a face into several modules where each of the modules contain different regions of facial features such as forehead, eyes, eyebrows, cheeks, nose and mouth. From the experiments, the region with the heaviest weight is the eyebrows, followed by mouth, eyes, cheeks, nose and forehead. Several works in face processing used optical flow, and that includes, face tracking [6], facial feature tracking [10] [14], facial expression classification [10] and face recognition [11]. [6] proposed a novel real-time face tracker which utilizes a modified version of the Viola-Jones algorithm for face detection. Forty-five uniform light condition videos from Boston Head Tracking Database are used in the experiment. Their proposed method easily outperformed the Viola-Jones face detector. A head tracker which is a combination of an optical flow and a template-based tracker is proposed in [10]. The tracking is robust as it is works well even the subject's face gets partially occluded. [10] also proposed two efficient approaches to track the mouth and eyebrow movements and this has allowed them to track and capture facial expressions. Normally, optical flow is used onto data with motion information. When a real-time face recognition involves varieties of facial expression, optical flow also has been applied. In [11] work, an integrated face recognition system that is robust against facial expressions by combining information from the computed intra person optical flow and the synthesized face image in a probabilistic framework, is proposed.

In the existing work of optical flow in face processing, the study of facial features' intensity with the optical flow has not been carried out yet. With the advantage of optical flow, the tracking of facial features' intensity will help to contribute to the automated recognition of the different facial expression intensity.

3. Our Work

In this study, video data from BU3DFE [7] database is used, which comprises of six basic facial expressions. BU3DFE database provides video of subjects with audio video interleaved (avi.) file format. All 90 videos (for 90 different subjects) display the sequences of facial expression in frontal face view. The facial expression video consists of approximately about 100 frames per second with a length of 4 seconds into 640 x 480 resolutions. The image sequences of BU3DFE database begin with the neutral state to the highest intensity level and then back to the neutral state of expression, refer to figure 1. The intensity of facial feature deformation is measured using optical flow.

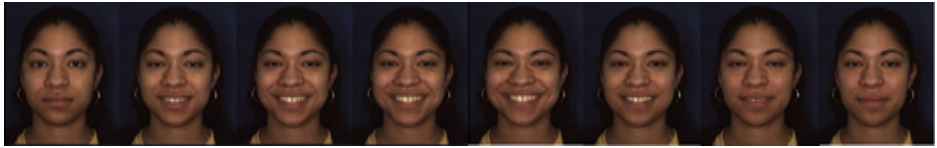

Figure 1: Facial Expression Sequences for Happy Expression [7]

Optical flow is the estimation of the apparent motion for an object between an observer and the scene [8]. The optical flow makes the motion measurement between two images frames possible. The method used is gradient based where it is based on local Taylor's series approximation to the image signal. The input for the optical flow is the images of a subject's face with the facial points. Each time the optical flow function is called, two video frames are extracted, which then will be used to measure the object's motion. To construct the optical flow between two images, equation 1, which is the optical flow constrained must be solved.

$$I_x u + I_y v + I_t = 0 \tag{1}$$

where:
$I_x$, $I_y$, and $I_t$ are the spatiotemporal image brightness derivatives
$u$ is the horizontal optical flow
$v$ is the vertical optical flow

At the initial phase, the video of facial expression from BU-3DFE database is converted to grayscale format. The conversion of RGB video into a grayscale format is to avoid the redundant pixel while extracting the information. Besides, the conversion into grayscale can simplify a large amount of the calculation in the next process [9]. Optical flow for the video is computed due to its ability to measure the motion between two video frames. Two video frames are used at one time, and one of the video frames is the original state of subject's faces, which is the neutral face and the other one is the non-neutral facial expression with a different level of expression intensity. By concentrating on a region of interest (ROI), help to observe the motion of the facial features. For example, to measure human task performance, ROI of gaze is used [12] [13] [15]. The segmentation process, i.e. segment the face into several ROIs, needs to be done to measure the magnitude of optical flow in order. The video frame is segmented into the 6X4 grid. As a result, there are three big regions: (1) eyes and eyebrow's region, (2) cheeks' region and (3) mouth region Figure 2 shows segmentation of subjects in this study.

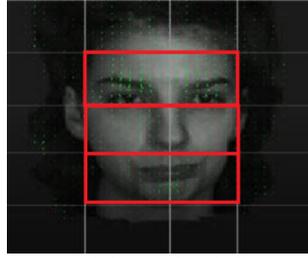

Figure 2: Segmented frame with optical flow. Image sourced from [7].

The magnitude of each facial feature is measured by computing the displacement of the initial position and previous position for each frame using Eq (2).

$$\bar{v} = \sqrt{(X_i - X)^2 + (Y_i + Y)^2} \qquad (2)$$

Finally, the facial expression in the video is analysed. The framework can be shown in the Figure 3.

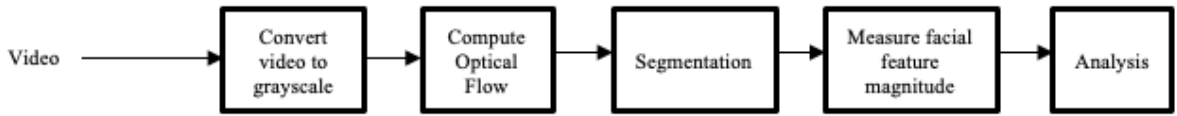

Figure 3: The facial expression analysis framework

4. Results and Analysis

In this section, the intensity of the facial features is analyzed where the magnitude of vector is calculated. The intensity of the facial feature versus sequence of frames is presented.

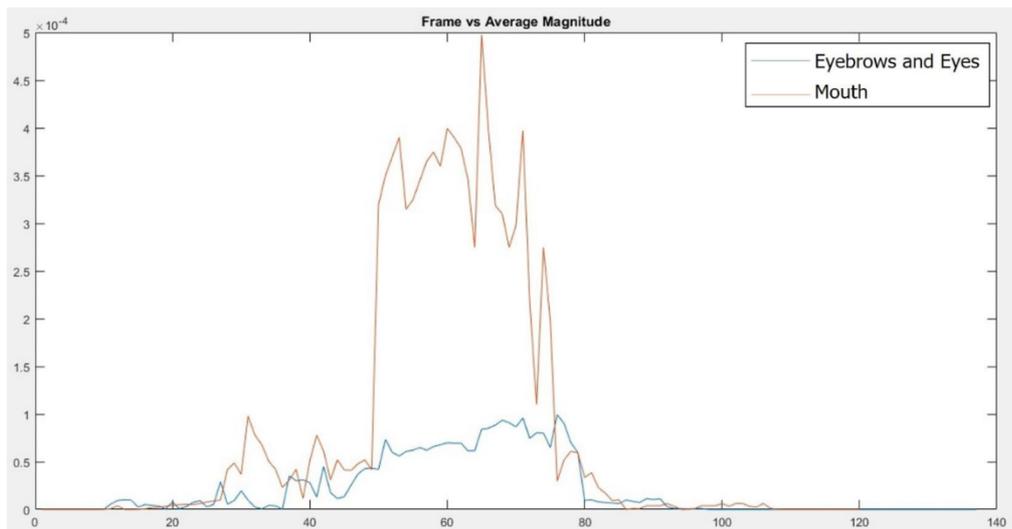

Figure 4: The facial features intensity of sad expression.

Figure 4 shows the sequence of frames versus average of magnitude of optical flow for sad expression. It shows that only eyes and mouth regions are deformed when sad expression occurred. There are no deformation happening in cheek region for sad expression is shown. In addition, it also shows that eyes region is deformed first before mouth. The intensity of mouth and eyes increasing concurrently. Then, the graph also shows that at certain frames, the intensity of the facial feature will decrease over the time where it means that the facial expression of the subject goes back to neutral state. In line with works in [17], the deformation of facial muscles for Sad expression consists of both eyebrows and lips. However, studies conducted in [4] which is based on facial feature distance could not capture this deformation as the distance for facial features involved in sad expression has a random mean and standard deviation results.

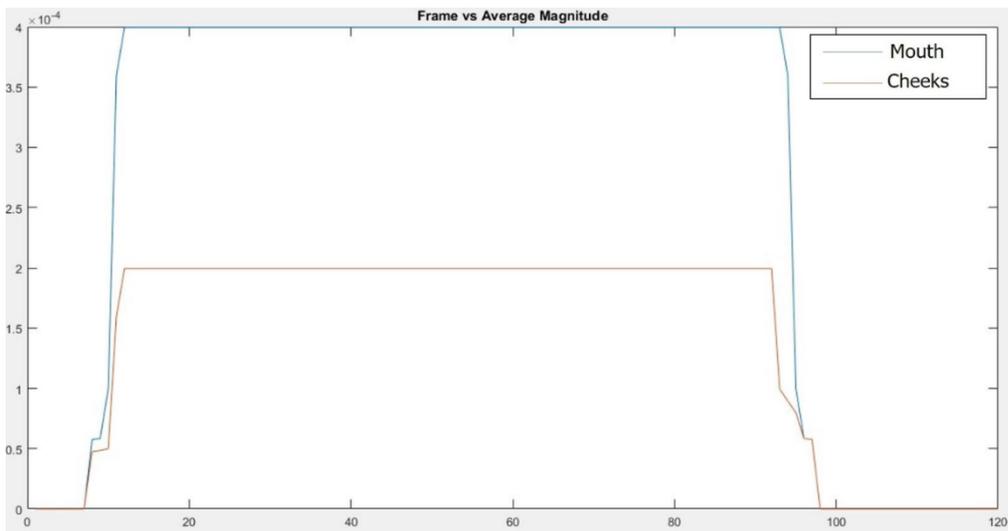

Figure 5: The facial features intensity of happy expression.

Figure 5 shows that only mouth and cheeks involve in happy expression. It also shows that mouth and cheeks segments are concurrently deformed together when happy expression are shown, and it started at frame 8. The mouth has a higher intensity rather than cheek. The intensity of facial features at both regions drop at frame 93.

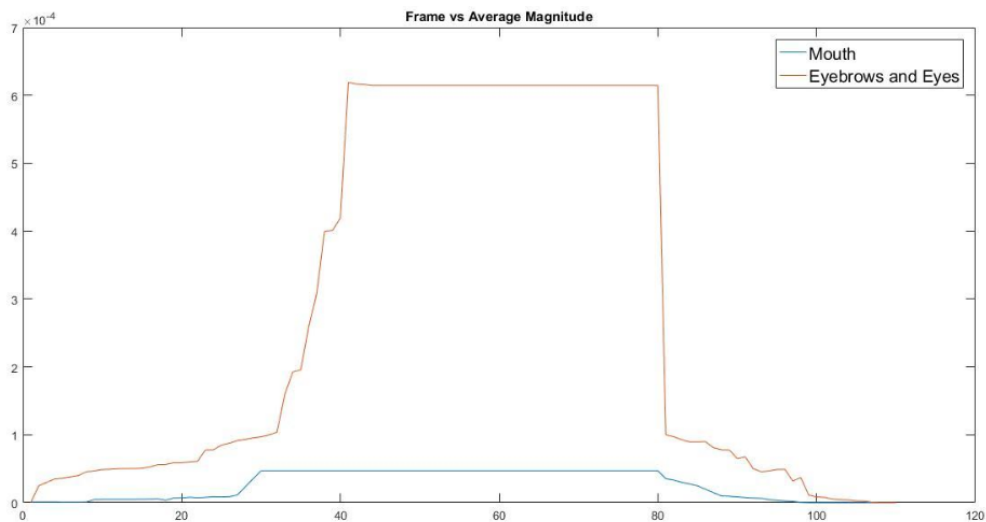

Figure 6: The facial features intensity of angry expression.

Figure 6 shows the facial features intensity of angry expression. From this graph, two regions have significant deformation when angry expression occurred, which are eyes and mouth region. It also shows that the eyes region has a higher intensity rather than mouth.

The intensity of facial features in both regions starts to increase and decrease at an approximately similar time.

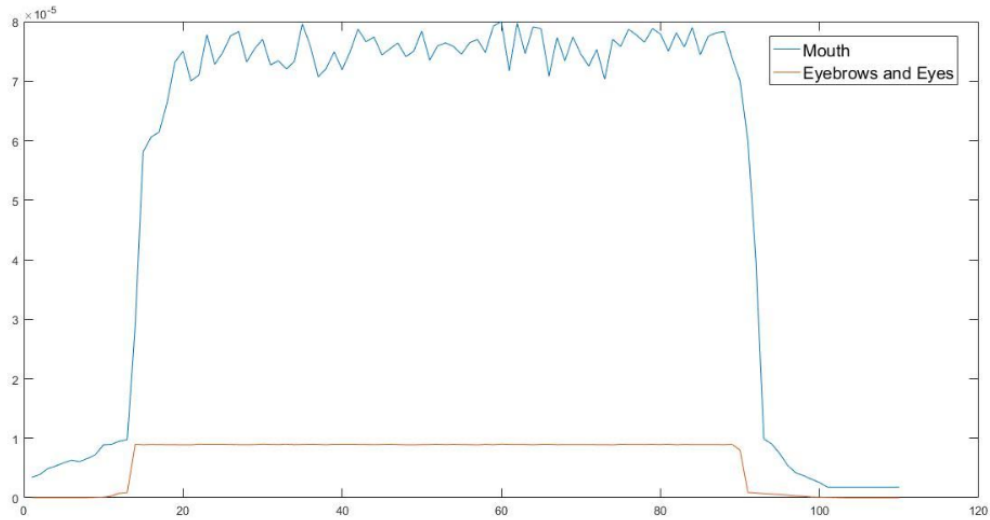

Figure 7: The facial features intensity of disgust expression.

Figure 7 shows the facial features intensity for disgust expression. The mouth region has the highest intensity rather than eyes' region. Based on the graph, it shows that the intensity of the facial features for both regions shows that they increase and decrease at an approximately similar time. However, at the highest intensity, the magnitude for eyes as well as the eyebrows is still compared to the mouth.

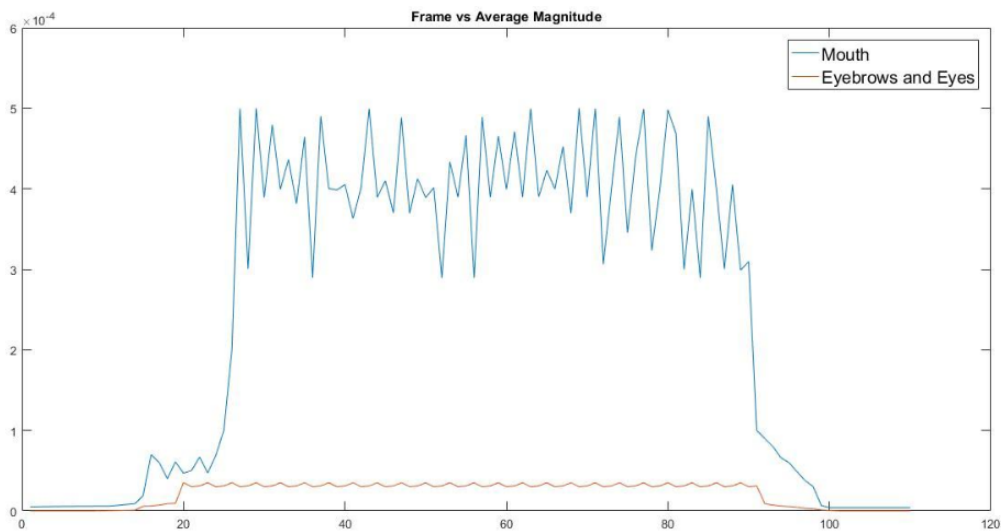

Figure8: The facial features intensity of fear expression

Figure 8 described the fear expression. The above graph shows that the mouth region has the highest intensity rather than the eyes' region. Mouth region deforms at the earliest frame compared with the eyes' region. The fluctuations of the mouth are significant across frames.

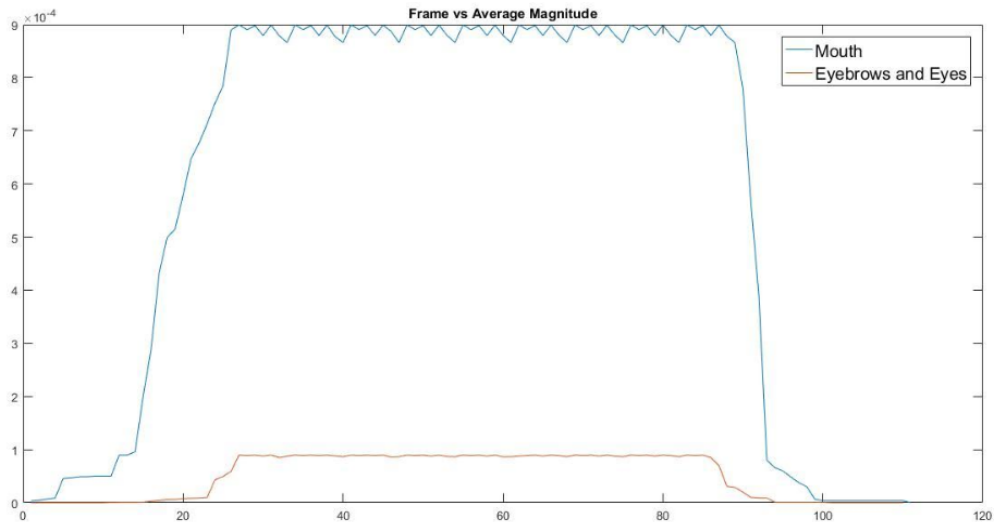
Figure 9: The facial features intensity of surprise expression

Figure 9 shows that the eyes and mouth regions are involved in surprise expression. Besides, it shows that the mouth region is more intense rather than the eyes' region during deformation. Mouth region also deforms first, and the facial features within the eyes' region deform a bit late and return to neutral state much earlier. According to [4], the significant facial muscle deformation for surprise expression is the mouth which is in line with our work.

For all expressions except happy, facial features in eyes and mouth region are significantly deformed when the emotion occurs. For happy expression, the facial features that involved are mouth and cheeks. The intensity within the mouth region is higher compared to the other facial features in all expressions except an angry expression. In addition, the intensity for mouth starts to deform first in angry, fear, surprise and disgust. For sad expression, the eyebrow and eyes deform at the earliest time frame. In the happy expression, both mouth and cheek are deformed together when emotion produced and based on the analysis, it shows that intensities of mouth are higher rather than cheeks.

5. Conclusion

In this study, optical flow is used to measure the motion of the facial expression by using Optical Flow Lucas Kanade algorithm. The magnitude of vector is measured to observe the facial features' intensities. In this study, the results supported the work in [4] that only certain facial features are deformed a particular facial expression is shown. Thus, it is not necessary to rely on all facial feature points in estimating facial expression intensity. However, results in [4] are different from our work.

Our next work is to combine other optical flow method which is Horn Schunck algorithm in our framework. The Horn Schunk algorithm is based on dense optical flow where we expect more interesting results. Also, in our future work, we will further our exploration in the pain assessment based on facial expression.